\pdfoutput=1

\documentclass[11pt]{article}

\usepackage[]{acl}

\usepackage{times}
\usepackage{latexsym}

\usepackage[T1]{fontenc}

\usepackage[utf8]{inputenc}

\usepackage{microtype}

\usepackage{graphicx}
\usepackage{amsmath}
\usepackage{mathtools}
\usepackage{amssymb}
\usepackage{float}
\DeclareMathOperator*{\argmax}{arg\,max}

\usepackage{nicefrac}

\usepackage{xcolor}
\definecolor{red}{RGB}{150, 0, 0}
\definecolor{green}{RGB}{0, 80, 10}
\definecolor{blue}{RGB}{25, 25, 150}
\newcommand{\red}{\color{red}}
\newcommand{\black}{\color{black}}
\newcommand{\green}{\color{green}}
\newcommand{\blue}{\color{blue}}
\usepackage{lipsum}

\title{Emergence of hierarchical reference systems in multi-agent communication}

\author{Xenia Ohmer \and Marko Duda \and Elia Bruni \\
        Osnabrück University\\
        \texttt{\{xenia.ohmer,mduda,elia.bruni\}@uos.de}}

\date{}

\begin{document}
\maketitle

\begin{abstract}
In natural language, referencing objects at different levels of specificity is a fundamental pragmatic mechanism for efficient communication in context. We develop a novel communication game, the \textit{hierarchical reference game}, to study the emergence of such reference systems in artificial agents. We consider a simplified world, in which concepts are abstractions over a set of primitive attributes (e.g., color, style, shape).
Depending on how many attributes are combined, concepts are more general (``circle'') or more specific (``red dotted circle''). Based on the context, the agents have to communicate at different levels of this hierarchy. Our results show that the agents learn to play the game successfully and can even generalize to novel concepts. To achieve abstraction, they use implicit (omitting irrelevant information) and explicit (indicating that attributes are irrelevant) strategies. In addition, the compositional structure underlying the concept hierarchy is reflected in the emergent protocols, indicating that the need to develop hierarchical reference systems supports the emergence of compositionality.
\end{abstract}

\section{Introduction}
\label{sec:introduction}

Humans excel at using language to convey information efficiently in context. A speaker does not have to communicate every detail. Rather, a listener can infer the intended meaning of an utterance by assuming that sufficient information was provided.
This idea was first explicitly formulated by \newcite{grice} in his conversational maxims, in particular the \textit{Maxim of Quantity}: ``1. Make your contribution as informative as is required (for the current purposes of the exchange). 2. Do not make your contribution more informative than is required.'' An illustration of this mechanism can be given in the form of a simple referential context. In a scene with a red circle and a green triangle, ``circle'' is enough information to identify the referent, whereas more complex scenes may require the speaker to name both object attributes---shape and color---to allow for an unambiguous interpretation. The Maxim of Quantity requires a hierarchical reference system, that allows the selection of the most appropriate level of specificity for a given context.

\begin{figure*}[htbp]
\begin{center}
\includegraphics[width=\textwidth]{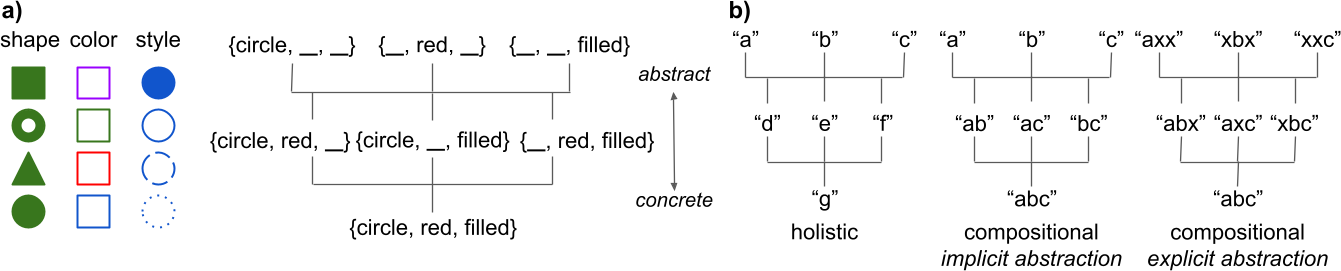}
\end{center}
\caption{\textbf{a)} Example of a concept hierarchy. Shown are all attribute values and the concept hierarchy constructed from the concept ``red filled circle''. \textbf{b)} Example languages for the concept hierarchy in part a). Possible abstraction strategies include holistic and compositional languages. In compositional languages, abstraction can further be indicated implicitly or explicitly.}
\label{fig:hierarchy}
\end{figure*}

In this paper, we follow the proposal by \newcite{higgins_2018} and define concepts as compositional abstractions over a set of primitive attributes (e.g., color, style, shape), see Figure \ref{fig:hierarchy}a. The concepts are maximally specific at the leaf nodes, where all attribute values are determined. Moving from the subordinate levels up to the superordinate levels, the number of concept-defining attribute values decreases. Thus, each parent concept is an abstraction (i.e. a subset) over its children and over the original set of attribute values. Given this definition of a concept hierarchy, we use a language emergence paradigm with artificial agents to study whether a corresponding reference system can emerge given a structured perception of the world. 

In most language emergence simulations, a sender and a receiver agent are trained on a reference game~\citep[e.g.,][]{havrylov_2017,lazaridou_2018,rodriguez-luna-etal-2020-internal,dagan-etal-2021-co}, based on the signaling game originally developed by \newcite{Lewis1969}. The sender sees a target object and sends a message to the receiver. Using that message, the receiver tries to identify the target among a set of distractor objects. Crucially, in the current form, reference games completely ignore that different contexts may require referential expressions at different levels of abstraction. Having no access to the distractors, the sender cannot choose relevant object attributes in a context-dependent way. Moreover, random sampling of the distractors typically encourages the sender to communicate all object attributes. Therefore, the standard reference game cannot account for the emergence of hierarchical concepts in communication.

We develop a \textit{hierarchical reference game} to address this shortcoming. Instead of an object, the sender receives a \textit{concept} as input. The concept is defined by an attribute vector (object) and a relevance vector (context). The relevance vector indicates for each attribute whether it is relevant in the current context or not. Based on the sender's message, the receiver must identify an object that instantiates the target concept among a set of distractors.
The input concepts have a compositional and hierarchical structure. While the game is designed to encourage communication at different levels of abstraction, it does not regulate how this abstraction is realized; in particular, there is no explicit pressure on the emergent language to reflect the compositional input structure. 


First, we evaluate if the agents can successfully play the game, i.e. communicate specific contextually relevant object attributes. Second, to measure whether the agents' strategies are systematic, we test whether they can generalize to novel concepts, and also whether they consistently use the same expressions for the same concepts at all levels of abstraction. Third, we investigate the emerging protocols to study the mechanisms by which systematic abstraction is achieved, see Figure \ref{fig:hierarchy}b. In natural language, there is \textit{holistic} abstraction as in ``Dalmatian'' $\subseteq$ ``dog'' $\subseteq$ ``animal''; but also \textit{compositional} abstraction as in ``filled red circle'' $\subseteq$ ``red circle'' $\subseteq$ ``circle''. Abstraction can further be \textit{implicit}, by omitting irrelevant attributes, and \textit{explicit}, by indicating that certain attributes are irrelevant (as in saying ``a circle of any color''). We evaluate which, if any, of these abstraction strategies are used by the agents.

Our work makes several contributions. We develop the hierarchical reference game and show that it can be used to model the emergence of referential expressions at different levels of abstraction. We also provide novel metrics to examine how the agents achieve abstraction. Working with different data sets, i.e. different concept hierarchies, allows us to disentangle data set specific and general effects. Not least, our results suggest that communication about concept hierarchies supports the emergence of compositionality.

\section{Related work}
\label{sec:related-work}

\paragraph{Referring expression generation (REG).} 
There has been a long history of research on understanding how people generate referring expressions, dating back to \citeauthor[]{winograd} (\citeyear{winograd}). The most influential work on REG in both the eighties \citep{appelt_1985, appelt_1987} and nineties \citep{reiter-dale_1992, dale_1995} integrated the Gricean maxims into their systems. The latter developed the \textit{iterative algorithm}, which was used and extended to model various aspects of REG \citep{krahmer_2012}.
Like \newcite[][]{dale_1995}, we define objects as sets of attribute-value pairs and consider the subset of referring expressions whose single purpose is to identify an object. 
However, our main focus is not to generate human-like referring expressions but rather to build artificial agents capable of hierarchical reference. Hence, we ignore many effects that play a role in human REG, for example basic level categories \citep{rosch_1975, rosch_1976}.

By now, several large-scale data sets of referring expressions for complex real-world images have been collected and are used to train deep neural networks (DNNs) \citep[e.g.,][]{kazemzadeh-etal-2014-referitgame, mao_2016, yu_2016, luo_2017, yu_2018, Luo_2020_CVPR}. The data sets are collected in a reference game setup: one participant has to refer to a target entity in a given image, and the other participant has to identify the target. Models are often trained on both components, expression generation and comprehension \citep[e.g.][]{mao_2016, luo_2017}.
Several REG models try to integrate deep learning with computational accounts of pragmatic reasoning \citep[e.g.,][]{monroe_2015, andreas-klein-2016-reasoning, le_2022}, such as the Rational Speech Act framework \citep{frank_2012_rsa}. Our model also implements expression generation (sender) and comprehension (receiver) using DNNs but hard codes pragmatic inferences in the relevance vector. Most importantly, the agents are not trained on a labeled data set but develop their own referring expressions in a language emergence game.

\paragraph{Emergent multi-agent communication.} 
Language emergence simulations are popular in evolutionary linguistics as well as AI research. In evolutionary linguistics, they are used to study the origins and evolution of human and animal communication \citep[e.g.,][]{cangelosi_2002, kirby_2002_overview, wagner_2003}. In AI research, they are used with the aim of building artificial agents capable of flexible and goal-directed language use, which arguably relies on grounding language in interaction \cite[e.g.,][]{steels_2001, steels_2003, lazaridou_2020}. 

Starting with \newcite[][]{foerster_2016} and \newcite[][]{lazaridou_2017}, there has been an increasing interest in language emergence simulations with DNN agents \citep[for a review, see][]{lazaridou_2020}. 
These approaches stand in contrast to the currently dominant DNN models in NLP, which learn passively by being exposed to large amounts of text \citep{bisk-etal-2020-experience}. As discussed above, in many implementations, hierarchical reference systems cannot emerge because the sender does not have access to information about the context. Even in the rare cases where it does \citep[e.g.,][]{lazaridou_2017, dessi_2021}, the emergence of hierarchical reference has not yet been investigated.


\section{General setup}
\label{sec:setup}

\subsection{Concept representation}

We use symbolic, disentangled input representations. A concept is composed of an object vector and a relevance vector. Objects have $n$ attributes and each attribute can take on $k$ values.\footnote{We present objects as $n$-hot encodings to the agents, such that each object $o \in \lbrace0,1\rbrace^{nk}$.} The relevance vector $r \in \lbrace0,1\rbrace^n$ indicates which attributes are relevant ($1$) and which ones are irrelevant ($0$). 
E.g., if the sender's input is $(4,3,1)(1, 0, 0)$, the concept in question is $(4,\:\_\:,\:\_\:) := \lbrace (4, x, y)\: | \:x, y\in \mathbb{N}, 1 \le x,y \le k\rbrace$. Object $(4, 3, 1)$ could instantiate the attributes \textit{shape}, \textit{color}, and \textit{style} with specific values such as \textit{circle}, \textit{red}, and \textit{filled} (see Figure \ref{fig:hierarchy}a).  Relevance vector $(1,0,0)$ would then indicate that only the first attribute value, \textit{circle}, is relevant and must be communicated. 

\subsection{Hierarchical reference game}

\begin{figure}[htbp]
\begin{center}
\includegraphics[width=\columnwidth]{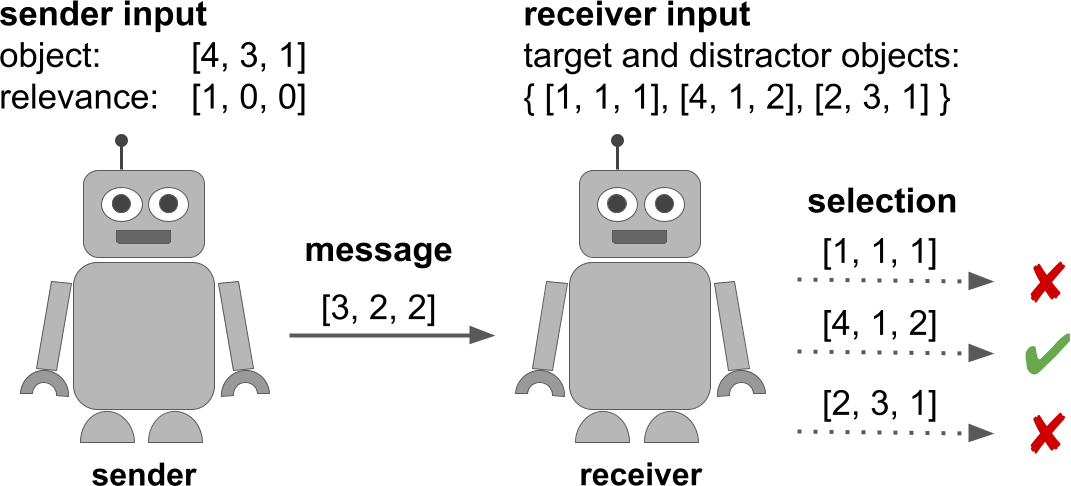}
\end{center}
\caption{Schematic illustration of the hierarchical reference game.} 
\label{fig:hierarchical_reference_game}
\end{figure}

Like the classical reference game, the hierarchical reference game is played by a sender, $S$, and a receiver, $R$. However, rather than communicating the input object as it is, the sender must abstract a concept from this object based on the relevance vector. One round of the hierarchical reference game proceeds as follows (see Figure \ref{fig:hierarchical_reference_game}):

\begin{enumerate}
    \itemsep-0.4em 
    \item An object, $o$, and a relevance vector, $r$, are sampled randomly and passed to $S$. 
    \item Based on this input, $S$ generates a message, $m$. The message is a concatenation of symbols from vocabulary $V$, $s_i \in V$, and has maximal length $L$, such that $m = (s_i)_{i\le L}$. 
    \item $R$ receives the message $m$, as well as a set of objects containing one target, $t$, and several distractors.
    $t$ has the same attribute values as the input object $o$ for relevant attributes (as defined by $r$), while the values of irrelevant attributes are sampled randomly. The distractors are constructed by sampling object instances of concepts that would arise from $o$ in combination with other relevance vectors than $r$. 
    \item Based on $m$, $R$ selects one object among target and distractors. 
\end{enumerate}

By our choice of distractors, we simulate an environment in which the relevance vector matches pragmatic needs: the speaker tries to be as specific as necessary in a given context. 
To further discourage communication of irrelevant attributes, we choose distractors that are more abstract than the target concept but still similar, by replacing exactly one 1 (\textit{relevant}) in the relevance vector with a 0 (\textit{irrelevant}). Additional experiments, where we sample distractors with equal probability from all levels of the concept hierarchy, can be found in Appendix \ref{appendix:control}.\footnote{In that case, the agents still learn to play the game successfully and to form abstract concepts but they have a stronger tendency to convey also irrelevant information.}

\subsection{Architecture}

Both agents are implemented as single-layer GRUs. The sender input is processed by two dense layers, one receiving the object vector and one the relevance vector, followed by a dense layer mapping a concatenation of these two representations to the sender's initial hidden state. The message is produced incrementally. At each time step, the sender generates a probability distribution over the vocabulary which is used to sample a symbol from the Gumbel-softmax distribution \citep{gumbel_2017}.
The GS distribution is a continuous distribution that approximates categorical samples, and whose parameter gradients can be easily computed via a reparameterization trick. 
The receiver processes the incoming message. An additional dense layer maps target and distractor objects onto embeddings. These embeddings have the same dimensionality as the receiver's hidden state. The receiver's selection probabilities are determined by applying a softmax function to the dot products between object embeddings and hidden state.

\section{Experiments}
\label{sec:experiments}

Our implementation is based on PyTorch, and uses the EGG toolkit \citep{kharitonov-etal-2019-egg}.\footnote{\url{https://github.com/facebookresearch/EGG}} Our code and results are available at \url{https://github.com/XeniaOhmer/hierarchical_reference_game}.

\subsection{Data sets}

In order to investigate how the number of attributes and the number of values per attribute influence the formation of abstract concepts, we use a set of different data sets, $D(n,k)\coloneqq \lbrace (m_1, \dots, m_n) \mid m_i \in \mathbb{N}_k\rbrace$ with $\mathbb{N}_k=\lbrace 1, ..., k\rbrace$ (see Table \ref{tab:input-size}). 

\begin{table}[htbp]
\begin{center}
\begin{tabular}{c|ccc}
 & $k=4$ & $k=8$ & $k=16$ \\
\hline
$n=3$ & $D(3,4)$ & $D(3, 8)$ & $D(3, 16)$ \\
$n=4$ & $D(4,4)$ & $D(4,8)$ & \\
$n=5$ & $D(5,4)$ \\
\end{tabular}
\end{center}
\caption{\label{tab:input-size} Input data sets with $n$ attributes and $k$ values. Data sets are labeled as $D(n,k)$.} 
\end{table}

We sample relevance vectors with equal probability from each level of the concept hierarchy.\footnote{If relevance vectors are sampled uniformly from the set of all relevance vectors, the amount of $0$ and $1$ entries follows a binomial distribution. Sampling relevance vectors with equal probability from each level of the hierarchy ensures that all abstraction levels occur equally often.}
We repeat that procedure until there are $10$ samples for each input object and each number of relevant attributes in the data set. In addition, we create $10$ distractors per sample. We reserve $20$\% of the data for zero-shot testing (see Section \ref{sec:evaluation}), and split the remaining data randomly into training and validation sets at a ratio of $0.75$/$0.25$.

\subsection{Hyperparameter selection and training}

In our simulations there is always exactly one target object for the receiver, i.e. only that object—and none of the distractors—is an instance of the target concept. The agents minimize the cross-entropy loss between target and selection. During training, a message is given by the GS distributions across symbols, whereas during testing the argmax values are used. Hence, it is possible to jointly update the weights of sender and receiver by backpropagating through the approximated ``discrete'' messages.

We conducted a hyperparameter search to identify model and training parameters leading to high performance on the validation set for the range of different data sets we use (for details see Appendix \ref{appendix:hyperparam}). Agents have an embedding layer with $128$ units and a hidden layer with $256$ units. The discrete messages are approximated using GS with an initial temperature of $1.5$, decaying exponentially at a rate of $0.99$. We train for $300$ epochs using Adam optimizer with batch size $32$ and learning rate $0.0005$. For all data sets, we use the number of attributes as maximal message length $L$. The minimal vocabulary sizes in Table \ref{tab:vocab-size} allow the sender to generate a distinct message for each input concept. They correspond to the number of attribute values plus one additional symbol to indicate irrelevance. The agents have an additional end-of-sequence symbol to terminate the messages before $L$ is reached. We run our experiments with a factor $f=3$ of the minimal vocabulary size. Additional experiments with other values for $f$ be found in Appendix \ref{appendix:control}.\footnote{Smaller factors make the task more difficult and performance decreases, while larger factors do not yield any further improvements.} We conduct $5$ runs per data set. 

\begin{table}[htbp]
\begin{center}
\begin{tabular}{c|ccc}
 & $k=4$ & $k=8$ & $k=16$ \\
\hline
$n=3$ & $5$ & $9$ & $17$ \\
$n=4$ & $5$ & $9$ & \\
$n=5$ & $5$ \\
\end{tabular}
\end{center}
\caption{\label{tab:vocab-size} Minimal vocab size for each data set. }
\end{table}

\subsection{Evaluation} \label{sec:evaluation}
We are interested in different aspects of the emergent language, and use existing as well as novel metrics to evaluate these. 

\paragraph{Zero-shot evaluation.} 
We generate two different zero-shot test sets. The first test set is used to evaluate whether the agents can generalize to novel \textit{objects}. It contains combinations of attribute values that do not occur in the training and validation data, and is reserved for testing after the data generation process. The second test set is used to evaluate whether the agents can generalize to novel \textit{abstractions}. We withhold abstractions from one value per attribute from the training data. The agents are trained from scratch on the remaining data and evaluated on the held-out data.

\paragraph{Message consistency and effectiveness.} 
To measure whether agents consistently use the same messages for the same concepts we employ information-theoretic metrics. Let $C$ be the set of concepts, and $M$ be the set of messages. The conditional entropy of messages given concepts, 
\begin{align*}
 \mathcal{H}(M\mid C) 
    &= -\sum_{c\in C,\, m\in M} p(c,m) \log \frac{p(c,m)}{p(c)} \,,
\end{align*}
measures how much uncertainty about the messages remains after knowing the concepts. Low values indicate that the agents consistently use the same messages for the same concepts, i.e. the language does not contain many synonymous expressions.  $\mathcal{H}(C\mid M)$, in turn, measures how much uncertainty about the concepts remains after knowing the messages and should therefore negatively correlate with the agents' performance. Low values indicate that agents effectively use messages that uniquely identify the target concept, i.e. the language does not contain many polysemous expressions. On this basis, we define a \textit{consistency} an \textit{effectiveness} score, using the marginal entropies $\mathcal{H}(C)$ and $\mathcal{H}(M)$ for normalization:
\begin{align*}
\text{consistency}(C,M) &= 1 - \frac{\mathcal{H}(M\mid C)}{\mathcal{H}(M)}\; \\ \text{effectiveness}(C,M) &= 1 - \frac{\mathcal{H}(C\mid M)}{\mathcal{H}(C)}\,.
\end{align*}
Finally, the normalized mutual information,
\begin{align*}
\mathcal{NI}(C,M) 
       &= \frac{\mathcal{I}(C,M)}{0.5\cdot \big( \mathcal{H}(C) + \mathcal{H}(M) \big)} \\
       &= \frac{\mathcal{H}(M) - \mathcal{H}(M\mid C)}{0.5\cdot \big( \mathcal{H}(C) + \mathcal{H}(M)\big)}\,,
\end{align*}
is a symmetric measure that combines the two conditional entropies into one score.

\paragraph{Symbol redundancy.} 
We develop this metric to approximate whether agents repeat information about attribute values in their messages. It assumes that each attribute value, $a_v$ (e.g $a$=\textit{color}, $a_v$=\textit{red}), is encoded by a specific symbol and counts how often that symbol is repeated given that $a_v$ is being encoded. The preferred symbol for each attribute value is defined $s^a_v \coloneqq \argmax_s \mathcal{I}(a_v, s)$, where we code for each message whether $s$ occurs at least once (the position of $s$ is irrelevant). Symbol redundancy is defined as the average number of occurrences of $s^a_v$ per message given that $a_v$ is part of the target concept.

\paragraph{Topographic similarity.}
Topographic similarity (\textit{topsim}) measures to what degree similar inputs are described by similar messages and is frequently used as a measure of compositionality. The metric calculates the pairwise distances between the inputs, as well as the pairwise distances between the corresponding messages, and then correlates the two distance vectors \citep{brighton_2006}. In the hierarchical reference game, we need to calculate the topographic similarity between messages and concepts. We use an $n$-hot encoding of the concepts ($n$ being the number of attributes) and treat abstraction from each attribute as an additional attribute value. If an attribute is relevant, that value is zero (no abstraction), if an attribute is irrelevant this value is one (abstraction) and overwrites the original attribute value. Assuming that each attribute can take on $k=4$ different values, the input encoding for the example in Figure \ref{fig:hierarchical_reference_game} becomes:
\begin{align*}
\text{sender input}&: 
[\, \blue4\, \red3\, \green 1\,\black]\, [\, 1\; 0\; 0\,]\, \text{(object + relevance)}\\
\text{encoding}&: [\,\blue0\, 0\, 0\, 1\, \black 0\, \red0\, 0\, 0\, 0\, \black 1\, \green0\, 0\, 0\, 0 \black 1\,]\,
\end{align*}
Analogously to \citet{lazaridou_2018}, we calculate the pairwise distances of the inputs using the cosine distance, and the pairwise distances between the messages using the edit distance. The topsim score is calculated as the Spearman correlation between the two resulting distance vectors. 

\paragraph{Disentanglement.} 
Positional disentanglement (\textit{posdis}) and bag-of-symbols disentanglement (\textit{bosdis})  are used to measure different types of compositionality \citep{chaabouni-etal-2020-compositionality}. For both metrics, concepts are encoded as for the topsim score. Posdis measures whether symbols in specific positions encode the values of a specific attribute, i.e. whether the compositional structure is order-dependent. Let $s_j$ be the $j$-th symbol of a message, then posdis is defined as
\begin{equation*}
    \text{posdis} = \frac{1}{L} \sum_{j=1}^{L} \frac{\mathcal{I}(s_j,a_1^j) - \mathcal{I}(s_j,a_2^j)}{\mathcal{H}(s)}\,,
\end{equation*}
where $L$ is the maximal message length, and $a_1^j$ and $a_2^j$ are the attributes that achieve the highest and second-highest mutual information with $s_j$ ($a_1^j = \argmax_a \mathcal{I}(s_j,a)$; $a_2^j = \argmax_{a\neq a_1^j} \mathcal{I}(s_j,a)$). Bosdis measures whether symbols refer to specific attribute values independent of their position. In that case, the language is permutation-invariant and only symbol counts matter. Let $n_j$ be a counter of the $j$-th symbol in a message, then bosdis is defined as 
\begin{equation*}
    \text{bosdis} = \frac{1}{|V|} \sum_{j=1}^{|V|} \frac{\mathcal{I}(n_j,a_1^j) - \mathcal{I}(n_j,a_2^j)}{\mathcal{H}(n_j)}\,,
\end{equation*}
where $V$ is the vocabulary size, and $a_1^j$ and $a_2^j$ achieve the highest and the second-highest mutual information with $n_j$.

\section{Results}
\label{sec:results}

In this section, quantitative and aggregated results will be presented. Random examples of concepts and messages, together with a qualitative analysis can be found in Appendix \ref{appendix:examples}. Appendix \ref{app:mappings} shows example mappings between abstract concepts and messages and Appendix \ref{app:abstraction-strategies} highlights different abstraction strategies.

\subsection{Performance and generalization}

\begin{figure}[htbp]
    \centering
    \includegraphics[width=\columnwidth]{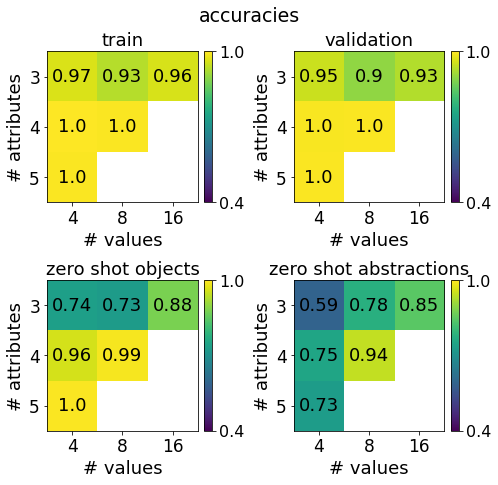}
    \caption{Mean accuracies across five runs for each of the training data sets. Shown are accuracies on the training set, the validation set, and the two zero-shot test sets.}
    \label{fig:accuracy}
\end{figure}

Figure \ref{fig:accuracy} shows the mean accuracies on training, validation, and zero-shot test sets for all data sets. Training accuracies (top left) and validation accuracies (top right) are very high for each data set, considering that chance performance is $<10\%$. 
Thus, the agents learn to refer to objects at different levels of abstraction, and their strategies do not overfit the training data. 

Accuracies for novel combinations of attribute values (bottom left) are consistently higher than accuracies for novel combinations of abstraction and attribute value (bottom right), except for $D(3,8)$. Accordingly, generalizing to novel abstractions of attribute values is harder than generalizing to novel objects. 
Both types of generalization tend to improve with the number of attributes as well as the number of values, which may be due to an increase in input space size \citep{chaabouni-etal-2020-compositionality}.
Similar to training and validation accuracies, generalization to novel objects reaches almost perfect accuracies, if there are many attributes.
While generalization to novel abstractions is more difficult, accuracies strongly exceed chance performance and are still very high for $D(3,16)$ with $84.76$\% and $D(4,8)$ with $94.38$\%. A large number of attribute values seems to be more important for generalizing to novel abstractions than for generalizing to novel objects, possibly because it is more useful to learn systematic abstraction if there are many attribute values. A strategy that abstracts from a certain attribute can be applied to more concepts if that attribute has many values (i.e. has more children on the concept hierarchy).  
Overall, the agents develop hierarchical reference systems and, with enough attributes and values, these systems generalize well to novel objects and novel abstractions.

\subsection{Mapping between concepts and messages}

\begin{figure}[htbp]
    \centering
    \includegraphics[width=\columnwidth]{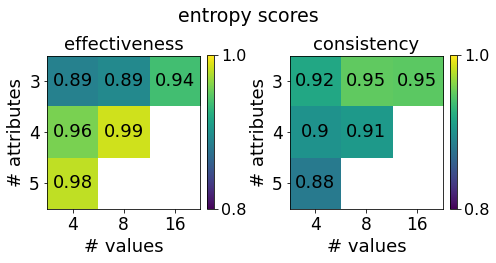}
    \caption{Mean effectiveness and consistency scores. We display the mean scores across five runs for each of the training data sets.}
    \label{fig:entropy}
\end{figure}

We determine the structure of correspondences between messages and concepts. Figure \ref{fig:entropy} shows the mean effectiveness and consistency scores.
The effectiveness score measures how much information about the target concept is contained in the message. It follows that the agents can only achieve high performance if the language is effective. The results show this interrelation, in that the pattern of effectiveness scores matches the pattern of training and validation accuracies across the different data sets. The consistency score, on the other hand, measures whether a concept is consistently mapped onto the same message, and high consistency is not necessary to achieve high performance. The score is higher for a larger number of attribute values, supporting the finding above that many values per attribute increase the pressure to develop systematic abstraction strategies. 

For each data set, the normalized mutual information lies between the effectiveness and the consistency score. It is generally high ($0.902 \leq \mathcal{NI} \le 0.945$), indicating that messages and concepts are strongly predictive of each other. A one-to-one correspondence between words and messages is not enforced by the setup because the message space is far larger than the concept space. The high entropy scores mean that a systematic mapping between concepts and messages, and therefore also systematic abstraction emerge nonetheless. 

\begin{figure}[htbp]
    \centering
    \includegraphics[width=0.95\columnwidth]{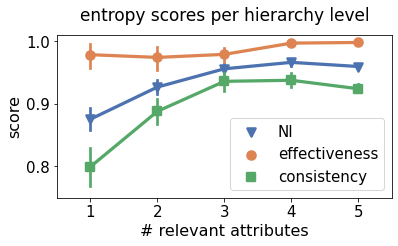}
   \caption{Mean entropy scores across all data sets for different numbers of relevant attributes: from left to right concepts become more concrete. Error bars indicate bootstrapped 95\% confidence intervals.}
    \label{fig:nsame_entropy}
\end{figure}

To analyze where the languages deviate from a one-to-one correspondence between concepts and messages, we consider the relation between entropy scores and level of abstraction (see Figure \ref{fig:nsame_entropy}). 
The mutual information between messages and concepts is higher for more concrete concepts. This effect is largely driven by an increase in consistency, while effectiveness is relatively constant across all levels of abstraction.
Thus, deviations from the one-to-one correspondence between concepts and messages occur mostly for abstract concepts. These deviations arise because different messages map to the same concept, not vice versa. In other words, the languages contain synonymy but no polysemy.

\subsection{Linguistic abstraction strategies}

Here, we look more closely at the types of internal message structures used to create a hierarchical reference system. 

\paragraph{Implicit versus explicit abstraction.}

\begin{figure}[htbp]
    \centering
    \includegraphics[width=0.49\columnwidth]{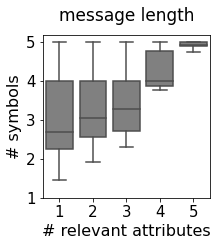}
    \includegraphics[width=0.49\columnwidth]{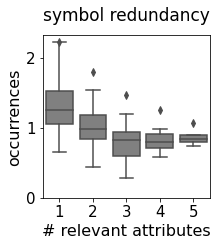}
   \caption{Average message length and symbol redundancy across data sets for different numbers of relevant attributes: from left to right concepts become more concrete.
   }
    \label{fig:implicit_abstraction}
\end{figure}

In natural language, there are implicit and explicit ways of communicating that attributes are irrelevant. A commonplace implicit strategy is to simply omit information about irrelevant attributes, e.g. one might say ``car'' rather than ``red car'' if sufficient. Since the maximal message length corresponds to the maximal number of relevant attributes, the agents could achieve a similar effect by using shorter messages for more abstract concepts or by using messages that contain more redundancies. Figure \ref{fig:implicit_abstraction} shows message length and symbol redundancy averaged across data sets for each level of abstraction. The agents indeed use implicit abstraction strategies and this is captured by both metrics. The message length decreases for more abstract concepts while symbol redundancy increases.
For abstract concepts, symbols that encode irrelevant attributes are either omitted or replaced by repetitions of symbols encoding relevant information. 

\begin{figure}[htbp]
    \centering
    \includegraphics[width=\columnwidth]{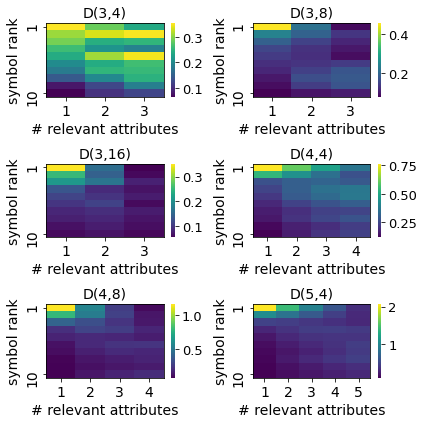}
   \caption{Average number of symbol occurrences per message for each level of abstraction. Symbols are ranked based on their occurrences for the most abstract concepts (i.e. with the fewest relevant attributes). Results are averaged across runs based on ranked symbols and shown only for the top ten ranks.}
    \label{fig:explicit_abstraction}
\end{figure}

Explicit abstraction would mean that the agents dedicate symbols to express that information is irrelevant. Such \textit{abstraction operators} should co-occur frequently with abstract concepts. We calculate the average number of symbol occurrences per message for each level of abstraction. We rank the symbols by their occurrences for the most abstract concepts to identify candidate symbols. Figure \ref{fig:explicit_abstraction} shows the results for the top ten candidates, averaging multiple runs for each ranked symbol. The ranking is visible in the left-most columns, where the number of occurrences per message decreases monotonously from the highest to the lowest rank. Strikingly, for all data sets except $D(3,4)$ only $1$--$3$ symbols occur very frequently together with very abstract concepts and the occurrence values decrease rapidly when going further down the ranks. Importantly, these symbols do not occur frequently at every level of the concept hierarchy. Rather, their usage decreases continuously as concepts become more concrete, as indicated by the gradient from left to right in the top rows. Thus, it seems likely that the agents use one or a few symbols to explicitly communicate information about the irrelevance of one or more attributes. The formation of abstraction operators is surprising since the message space is large enough to encode irrelevance differently, for example by combining symbols or using different symbols for different attributes. 

\paragraph{Compositional versus holistic abstraction.}

The hierarchical reference game requires the agents to repeatedly communicate the same attribute values but for different concepts---different because of the values of other attributes (traversing the hierarchy horizontally) or because of the level of abstraction (traversing the hierarchy vertically). Although the agents could develop holistic protocols, this repeated reference across contexts might encourage them to develop ``reusable'' mappings from attribute values to symbols, i.e. compositional expressions.

We use the different compositionality metrics to quantify the degree and nature of compositionality in the messages. Mean scores for each metric and data set can be found in Appendix \ref{appendix:compositionality}. The mean topsim score across data sets is $0.424$. The score is even higher, with $0.501$, if only concrete concepts are taken into account (as in a standard reference game). The mean posdis score across data sets is $0.115$ and the mean bosdis score $0.406$. So, there is compositional structure in the messages, and the agents prefer to use specific symbols per attribute value, independent of their position in the messages. 

\begin{figure}[htb]
    \centering
    \includegraphics[width=\columnwidth]{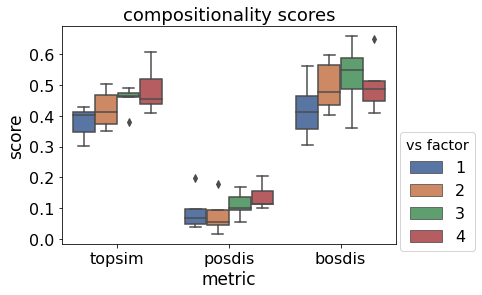}
    \caption{Boxplots of the compositionality scores for $D(4,8)$ and different vocabulary sizes.}
    \label{fig:compositionality_vs}
\end{figure}

In additional experiments (see Appendix \ref{appendix:control}), we trained the agents on $D(4,8)$ with different vocabulary sizes, using factors $f \in \lbrace 1, 2, 3, 4\rbrace$ of the minimal vocab size in Table \ref{tab:vocab-size}. While a fully positional encoding can be achieved with a smaller vocabulary ($f=1)$, a fully position-independent encoding requires a larger vocabulary. Mean training accuracies for $f=1$ are $0.936$, and for all other factors $> 0.99$. Figure \ref{fig:compositionality_vs} shows the compositionality scores for each factor. Surprisingly, all scores tend to increase with the vocabulary size, regardless of whether the corresponding type of compositionality requires a large vocabulary size or not. Usually, vocabulary size is reduced to increase the pressure for compositional solutions \citep{kottur-etal-2017-natural}. In our case, compositionality probably increases with vocabulary size because the emerging compositional structure is largely non-positional. 

\section{Conclusion}
\label{sec:conclusion}

In this work, we developed a hierarchical reference game to study the emergence of hierarchical reference systems. In the game, concepts are defined as abstractions over a set of attributes. To refer to these concepts, our agents developed abstract terms and used these terms systematically, in the sense that they could generalize to novel objects and novel abstractions. It seems that, aside from more obvious strategies such as leaving out irrelevant information, the agents developed abstraction operators to explicitly indicate the irrelevance of certain attributes. Even more surprisingly, for some data sets, they used \textit{the same} few symbols to indicate irrelevance across attributes, rather than a dedicated symbol per attribute. While the game design encourages the emergence of abstract concepts, the use of specific abstraction operators emerged without any explicit pressure. 

In addition, our results suggest that compositional language may emerge as part of a hierarchical reference strategy. In the classical reference game, the sender typically tries to communicate the union of all object attributes. Without additional pressures, the emerging languages are not compositional \citep{kottur-etal-2017-natural, van-der-wal-etal-2020-grammar, dagan-etal-2021-co}. In the hierarchical reference game, in contrast, the sender must pick out specific attributes for communication, which potentially stimulates disentanglement. This interpretation is in line with the finding that the emergence of compositionality is supported by an increasing number of relevant events that can be referred to \citep{Nowak2000}. In the hierarchical reference game, cross-situational reuse is increased, as reference to attribute values occurs not only across objects but also across levels of abstraction.

We envision two main directions for future work. First, we would like to implement a hierarchical reference game with raw visual inputs instead of symbolic input vectors. 
\citet{higgins_2018} have developed a neural network (SCAN) that not only learns disentangled visual primitives in an unsupervised manner but also abstractions over such primitives from very few symbol-image pairs that apply to a particular concept. Combining our language emergence game with such a network would allow us to study the simultaneous emergence of abstract visual and linguistic concepts, as well as interactions between these two processes. Second, instead of hard-coding the relevance vector, the relevance of certain attributes should arise from the agents' intentions. 
Ideally, the agents would play a more complex game and determine themselves which properties of the environment are relevant for their objectives in the current context. Besides, sender and receiver could use pragmatic reasoning \citep[as for example in][]{choi_2018,kang_2020,yuan_2020} to encode and decode which attributes should be emphasized to communicate certain concepts.

\section*{Acknowledgements}

This work was funded by the Deutsche Forschungsgemeinschaft (DFG, German Research Foundation)---GRK 2340. We would like to thank Dieuwke Hupkes, Leon Schmid, Lucas Weber, and the anonymous reviewers for their valuable feedback and suggestions.


\bibliography{anthology,custom}
\bibliographystyle{acl_natbib}


\newpage
\appendix

\section{Varying distractor sampling and vocab size}
\label{appendix:control}

\subsection{Setup}

We conduct control experiments, changing the vocabulary size, and changing the distractor sampling strategy. In the original experiments, the message space is much larger than the space of concepts that need to be communicated. By reducing the vocabulary size, we aim to test whether a smaller message space increases the probability of one-to-one associations between concepts and messages. In addition, the distractors are sampled from concepts that are one level more abstract than the target concept on the concept hierarchy. Here, we relax this assumption by sampling distractors from all levels of the concept hierarchy with equal probability. Together, these additional experiments allow us to extend our results to different vocabulary sizes, and more general distractor distributions. 

We focus on a single data set. We use the data set with four attributes and eight values per attribute, $D(4,8)$, which achieved the highest mean validation accuracies and normalized mutual information scores in the original setup. In the original experiments, we used a factor of $3$ of the minimal vocabulary size $9$ ($8$ for each value plus $1$ for coding irrelevance). Now, we run the same experiment for factors of $1$, $2$, and $4$; and in addition, we repeat the experiment for each factor with the alternative sampling strategy. Again, we conduct five runs for each factor and sampling strategy. 

\subsection{Results}

\begin{figure}[htb]
    \centering
    \includegraphics[width=0.9\columnwidth]{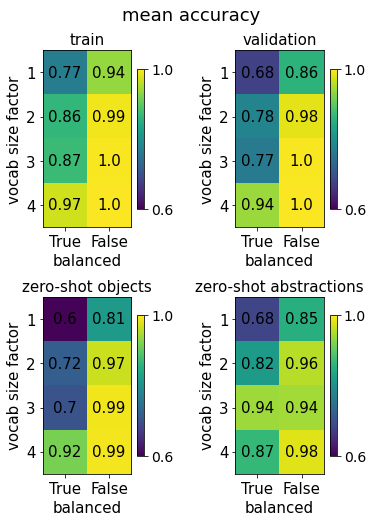}
    \caption{Mean accuracies for the control experiments across five runs, on the training data, the validation data, and the two zero-shot test sets. The $y$-axis gives the factor used to determine the vocabulary size, vocab size = factor $\times$ minimal vocab size, and the $x$-axis indicates whether distractors are sampled from concepts that are one level more abstract than the target concept (unbalanced), or sampled from all levels of the concept hierarchy with equal probability (balanced).}
    \label{fig:accuracy_control}
\end{figure}

Figure \ref{fig:accuracy_control} shows the accuracy scores for the different vocabulary size factors, and the different distractor sampling strategies, where \textit{unbalanced} refers to the original strategy of selecting distractors from more abstract concepts, and \textit{balanced} refers to the control strategy of sampling distractors with equal probability from all levels of abstraction. For both sampling strategies, performance is higher if the vocabulary size is large, likely because having a larger message space increases the number of solutions. A larger vocabulary size seems to be particularly important if distractor sampling is balanced. 

The original, unbalanced sampling strategy achieves higher performance than the control strategy on all data sets. So, choosing distractors very similar to the target facilitates learning, and probably also abstraction as suggested by the zero-shot evaluation with new abstractions. To make sure that the unbalanced sampling strategy only facilitates learning but does not make the task easier, we run an ablation test. We evaluate each sender-receiver pair on the validation set of the sampling strategy that was \textit{not} used for training. For all vocabulary sizes and runs, the agents perform better on the balanced validation set compared to the unbalanced validation set, regardless of the sampling method used during training. In conclusion, sampling distractor concepts that are very similar to the target concept makes the task more difficult but improves learning by increasing the pressure to communicate only relevant aspects, and thereby to develop abstract concepts.

\begin{figure}[H]
    \centering
    \includegraphics[width=0.75\columnwidth]{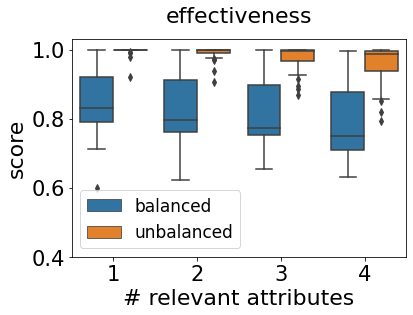}\\
    \vspace{0.3cm}
    \includegraphics[width=0.75\columnwidth]{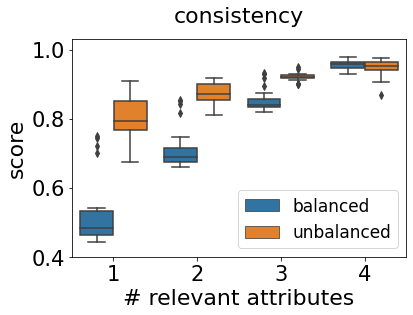}
    \caption{Effectiveness and consistency scores for balanced and unbalanced distractor sampling, separated for each level of abstraction. Distributions show the results across the different vocabulary sizes and runs. The level of abstraction is given on the $x$-axis: from left to right the concepts become more concrete.}
    \label{fig:entropy_control}
\end{figure}

These results are confirmed by the entropy-based evaluation metrics shown in Figure \ref{fig:entropy_control}. Effectiveness and consistency are consistently lower for the balanced distractor sampling strategy. However, while the level of abstraction does not have a strong effect on the difference in effectiveness scores, the difference in consistency scores decreases continuously with the level of specificity. In line with the generalization ability, this suggests that the unbalanced sampling strategy supports the formation of abstract concepts by reducing the probability of successful target selection if irrelevant attributes are communicated.

\section{Hyperparameter search} \label{appendix:hyperparam}

We ran our hyperparameter search for the three data sets spanning up the space of all data sets we use, $D(3,4)$, $D(5,4)$, and $D(3,16)$ (see Table \ref{tab:input-size}). We expected that hyperparameters working across all these extreme cases should also work for interpolations between them. Certain hyperparameters were fixed across the search. We used GRUs with Adam optimizer, and a GS temperature of $1.5$ with an exponential decay rate. Message length cost was $0$, and vocab size factor 3. We varied the following hyperparameters: 
\begin{itemize}
    \item batch size: $\lbrace 32, 64, 128\rbrace$
    \item learning rate: $\lbrace 0.0005, 0.001\rbrace$
    \item hidden layer dimension: $\lbrace 128, 256\rbrace$
    \item embedding layer dimension: \\always half of the hidden layer dimension
    \item GS temperature decay rate:  $\lbrace 0.97, 0.99\rbrace$
\end{itemize}
\noindent For the grid search we stopped the training process after $60$ epochs. All results can be found in our repository.

\section{Compositionality scores} \label{appendix:compositionality}

\begin{figure}[H]
    \centering
    \includegraphics[width=\columnwidth]{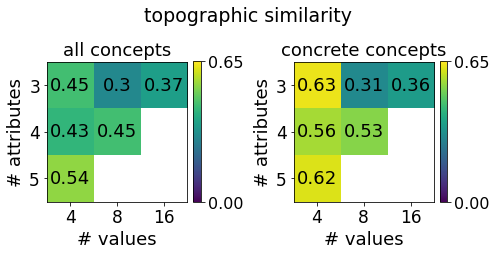}
    \includegraphics[width=\columnwidth]{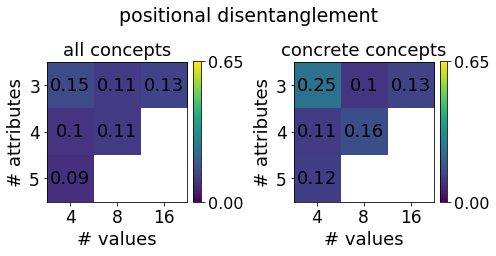}
    \includegraphics[width=\columnwidth]{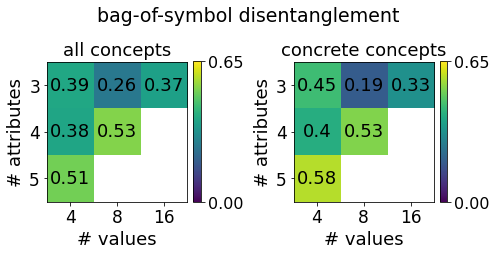}
    \caption{Mean compositionality scores per data set. }
    \label{fig:my_label}
\end{figure}

\section{Qualitative examples} \label{appendix:examples}

This section provides qualitative examples of concept-message pairs. Examples were randomly selected from the first run of each data set. Interestingly, this microcosm of random examples reflects all communication patterns that were identified in the quantitative analyses.

\subsection{Mappings between concepts and messages}
\label{app:mappings}

\begin{figure*}[ht!]
    \begin{center}
    \includegraphics[width=\textwidth]{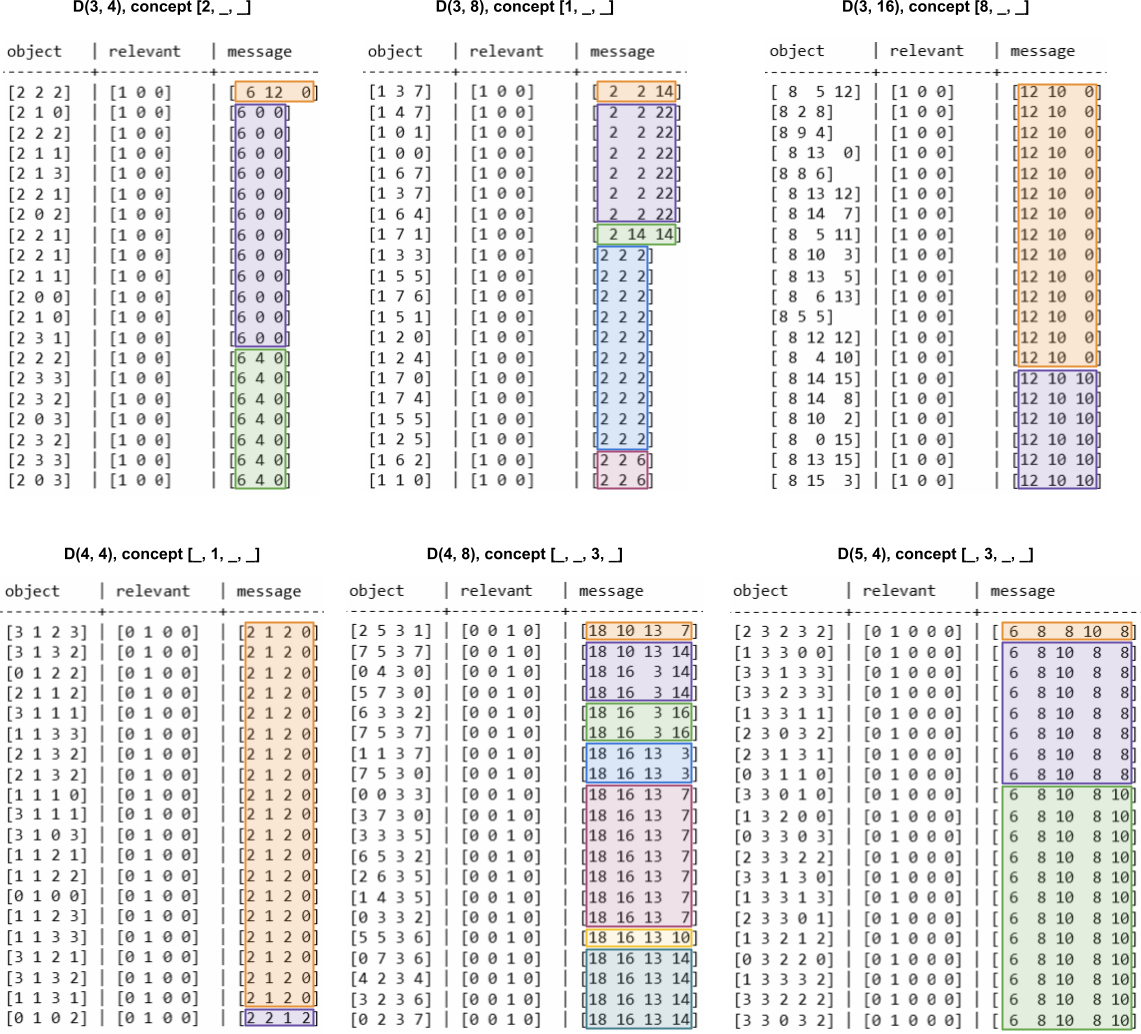}
    \end{center}
    \caption{Example messages for one abstract concept per data set. For each data set, we randomly select a concept at the highest level of abstraction. We then randomly select $20$ instances of that concept in the training data and display these instances together with the corresponding messages (from the first run). The same messages are grouped together in colored boxes.}
    \label{fig:concepts}
\end{figure*}

We are interested in whether the agents use the same message to refer to abstract concepts regardless of how these concepts are instantiated. Figure \ref{fig:concepts} shows the messages for a randomly selected concept at the highest level of abstraction (only one attribute is relevant), instantiated by different attribute vectors for each data set. Shown are twenty randomly selected instances each, and the examples are sorted by message. 

Abstraction is relatively systematic. For all data sets, the agents group together different concept instances in their messages. For some data sets, the instances are grouped under very few messages. For example, the sender trained on $D(4,4)$ groups together all example instances of the concept $(\_,1,\_,\_)$ under just two different messages ($(2, 1, 2, 0)$ and $(2, 2, 1, 2)$). Across data sets, $2$, $3$, $5$, or $7$ different messages are used to describe the $20$ example instances. In line with the quantitative results, there is no perfect one-to-one correspondence between \textit{abstract} concepts and messages. How many different messages are used also depends on the abstraction strategy (see Appendix \ref{app:abstraction-strategies}).

\subsection{Abstraction strategies}
\label{app:abstraction-strategies}

\begin{figure*}[ht!]
    \begin{center}
    \includegraphics[width=0.9\textwidth]{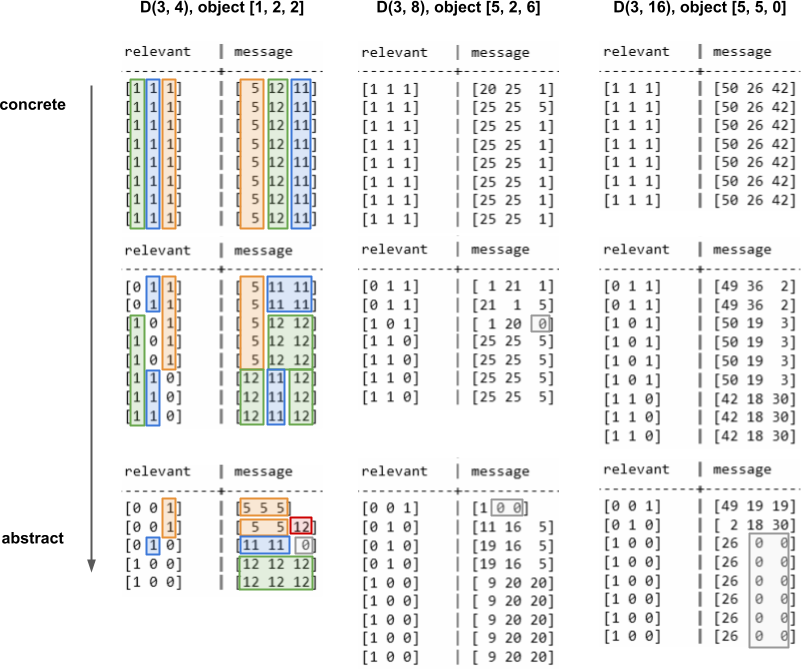}
    \end{center}
    \caption{Messages for a random object at each level of abstraction available in the training data. The corresponding messages are shown for the first run of each data set: $D(3,4)$, $D(3, 8)$, and $D(3, 16)$. The highlighted patterns are explained in the text.}
    \label{fig:qualitative1}
\end{figure*}

\begin{figure*}[ht!]
    \begin{center}
    \includegraphics[width=\textwidth]{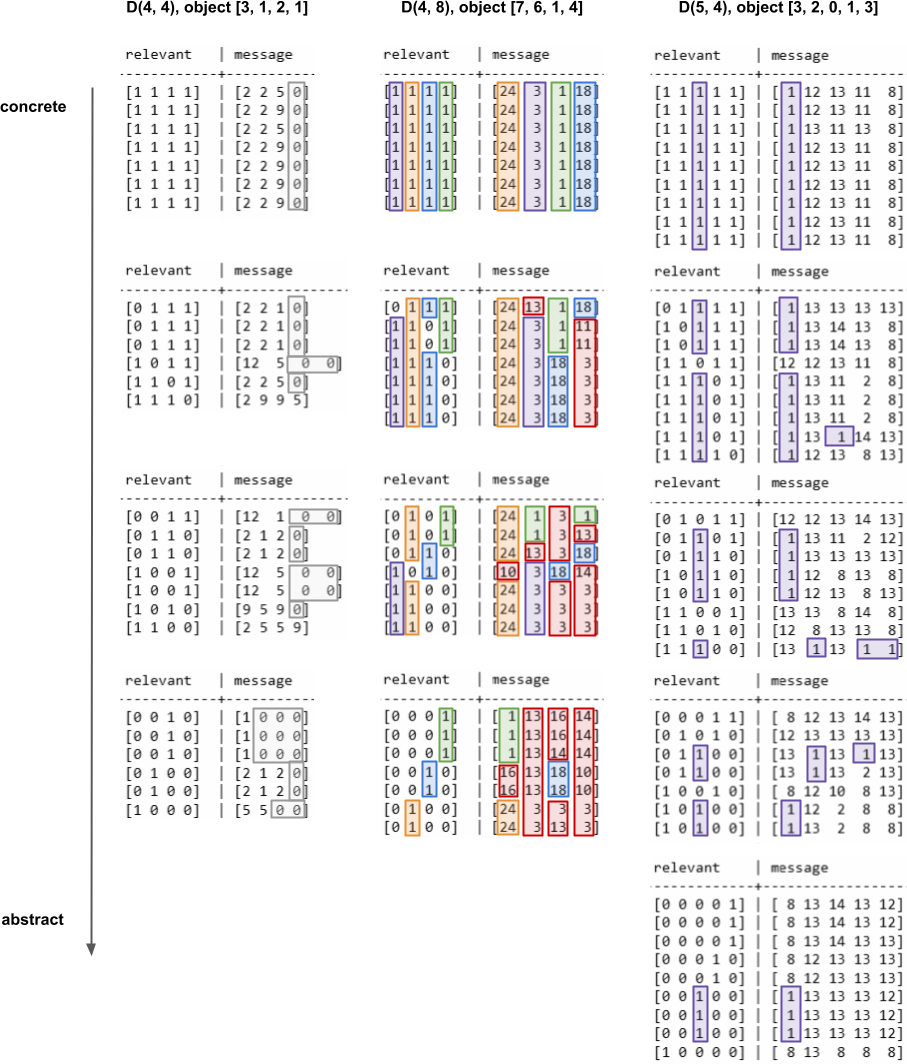}
    \end{center}
    \caption{Messages for a random object at each level of abstraction available in the training data. The corresponding messages are shown for the first run of each data set: $D(4,4)$, $D(4, 8)$, and $D(5,4)$. The highlighted patterns are explained in the text.}
    \label{fig:qualitative2}
\end{figure*}

To visualize the agents' abstraction strategies, we randomly selected an object (i.e. attribute vector) for each data set, and show the messages for that object across the concept hierarchy, so for each abstraction in the training set. Because of their large number the examples are split into two figures, Figure \ref{fig:qualitative1} ($D(3,4)$, $D(3, 8)$, and $D(3, 16)$) and Figure \ref{fig:qualitative2} ($D(4,4)$, $D(4, 8)$, and $D(5,4)$), which will be analyzed together. The messages will first be analyzed for compositional structure, and then for implicit versus explicit abstraction.

\paragraph{Compositional versus holistic abstraction.}

For some data sets, the agents seem to use \textit{trivially compositional} messages, i.e. messages whose meaning corresponds to the intersection of meanings of their constituents. An unambiguous pattern can be identified for $D(3,4)$ and $D(4,8)$, where a mapping between each attribute value and a specific symbol can be established (color-coded in orange, green, blue, and purple). In the case of data set $D(4,8)$, the number of additional ``filler'' symbols increases with the level of abstraction (color-coded in red). These might serve as abstraction operators (see below). For other data sets, like $D(5,4)$, such mappings can only be identified for specific attribute values (color-coded in purple). Here, symbol $1$ occurs if and only if the third attribute has the value $3$. For all identified mappings, the symbols are used to encode specific attribute values relatively independent of their position, which is in line with the high bosdis and low posdis scores in our quantitative analyses. 

For the remaining data sets, the messages are not unstructured but no one-to-one correspondences can be identified. For example, looking at the messages for $D(3,16)$, symbol $26$ might encode value $5$ at position $1$ for the most concrete and most abstract concepts but is not used at the intermediate level of abstraction. So, while the abstraction strategies are almost perfectly compositional in some cases, there are large variations between data sets, and potentially also runs and concepts. 

\paragraph{Implicit versus explicit abstraction.} The examples show instances of implicit and explicit abstraction strategies. Implicit abstraction is identified through shorter messages and more symbol redundancy for higher levels of abstraction; explicit abstraction through the use of abstraction operators. At least for some data sets, the messages tend to become shorter with increasing abstraction. E.g. messages become shorter in the case of $D(3,16)$ and $D(4,4)$ (the end-of-sequence symbol $0$ is color-coded in gray). 

Symbol redundancy and abstraction operators can best be identified in reference systems with compositional structure. $D(3,4)$ is a perfect example of increasing symbol redundancy. Each symbol corresponds to a specific attribute value, and symbols are repeated to fill up the messages for more abstract concepts. E.g., the concept (\color{green}1\color{black}, \color{blue}2\color{black}, \color{orange}2\color{black}) is encoded as (\color{orange}5\color{black}, \color{green}12\color{black}, \color{blue}11\color{black}), the concept (\color{green}1\color{black}, \_, \color{orange}2\color{black}) as (\color{orange}5\color{black}, \color{green}12\color{black}, \color{green}12\color{black}), and the concept (\color{green}1\color{black}, \_, \_) as (\color{green}12\color{black}, \color{green}12\color{black}, \color{green}12\color{black}). 

\begin{figure}[ht!]
    \begin{center}
    \includegraphics[width=0.75\columnwidth]{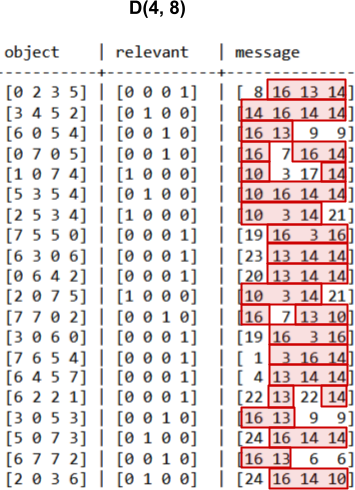}
    \end{center}
    \caption{Messages for $20$ randomly selected concepts at the highest level of abstraction, for the first run of $D(4,8)$. The highlighted patterns are explained in the text.}
    \label{fig:qualitative3}
\end{figure}

$D(4, 8)$, on the other hand, is a perfect example of explicit abstraction. As messages become more abstract the frequency of symbols that do not encode an attribute value ($\lbrace 3, 10, 11, 13, 14, 16\rbrace$, marked in red) increases. Note that symbol $3$ seems to serve both roles, encoding an attribute value as well as encoding abstractions. To confirm the intuition that these additional symbols serve as abstraction operators, we look at other abstract concepts for $D(4,8)$. Figure \ref{fig:qualitative3} shows the messages for $20$ random examples. Indeed, at least two of the abstraction operators occur in each message. Only symbol $11$ does not occur and might serve a different function. The quantitative analyses suggest that usually less abstraction operators are used than in this specific example. Less compositional protocols may also use explicit abstraction operators or symbol redundancy but these cannot easily be identified in a qualitative analysis.

\end{document}